\documentclass[11pt]{article}

\usepackage{acl}

\usepackage{times}
\usepackage{latexsym}

\usepackage[T1]{fontenc}

\usepackage[utf8]{inputenc}

\usepackage{microtype}

\usepackage{inconsolata}

\usepackage{graphicx}
\usepackage{booktabs}
\usepackage{tabularx}
\usepackage{soul,color}
\usepackage{amsmath}

\title{From Triage to Impact: A Multi-Hop Reasoning Benchmark for Explainable Persian Hate Speech}

\author{
  Ghazal Kalhor\textsuperscript{1} \quad
  Sima Kamandloo\textsuperscript{2} \quad
  Mehrsa Shamsian\textsuperscript{3} \quad
  Behnam Bahrak\textsuperscript{4}
  \\
  \textsuperscript{1}EPFL \quad
  \textsuperscript{2}Shahid Beheshti University \\
  \textsuperscript{3}University of Social Welfare and Rehabilitation Sciences \quad
  \textsuperscript{4}Khatam University
  \\
  \small{
    \textbf{Correspondence:}
    \href{mailto:ghazal.kalhor@epfl.ch}{ghazal.kalhor@epfl.ch},
    \href{mailto:b.bahrak@teias.institute}{b.bahrak@teias.institute}
  }
}

\begin{document}

\maketitle
\begin{abstract}
Hate speech has become increasingly prevalent on social media, raising concerns about its detection, interpretation, and moderation. Persian, a relatively low-resource language, lacks sufficient annotated data and computational resources for fine-grained hate speech analysis. In this study, we introduce a fine-grained, multi-hop benchmark for explainable Persian hate speech analysis, covering content triage, target identification, hate tactic and moral foundation analysis, textual evidence extraction, harmful narrative inference, and severity assessment. By evaluating state-of-the-art large language models (LLMs), we find that performance generally declines as tasks require more fine-grained semantic and inferential reasoning. Proprietary models generally outperform open-weight models, while reasoning-oriented models show no consistent advantage. Incorporating intermediate reasoning steps substantially improves severity prediction compared with one-shot assessment. Our analysis further reveals that \emph{Incitement to Violence or Harm} is most strongly associated with higher severity, while political-affiliation targets are particularly associated with subversion-based moral framing. Overall, this work provides insights into LLM-based hate speech analysis and supports more accountable moderation of Persian-language social media. The benchmark is publicly available at \url{https://github.com/kalhorghazal/From-Triage-to-Impact}.
\end{abstract}

\begin{figure*}[t]
  \centering
  \includegraphics[width=\linewidth]{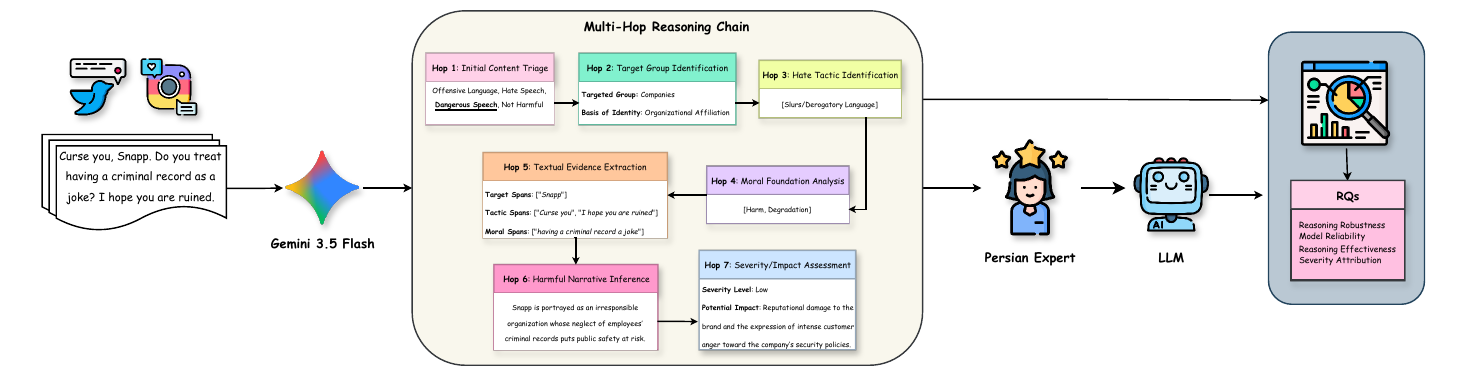}
  \caption{Pipeline illustrating multi-hop reasoning steps, human verification, and LLM evaluation.}
  \label{fig:pipeline}
\end{figure*}

\section{Introduction}

From the advent of online social media, hate speech\footnote{This paper includes potentially offensive or violent hate speech examples.} has become an increasingly prevalent and harmful behavior among users of these platforms. It expresses hatred toward a targeted group or seeks to demean or insult its members, undermining their dignity, inclusion, and well-being \cite{davidson2017automated, waldron2012harm}. The Persian language is used across multiple geopolitical contexts, from Iran to the diaspora, and is associated with real-world stakes including migration tensions, religious marginalization, and gender and political polarization, making it a high-impact language for NLP-based content moderation \cite{bokaei2025culture}.

Despite the societal importance of Persian hate speech, the language remains low-resource in this domain, as most existing hate speech detection systems are English-centric \cite{trager2025mftcxplain, tonneau2024languages, yang2023hare, poletto2021resources}. Persian-focused approaches often rely on black-box classification with binary or coarse-grained labels that lack explainability \cite{kebriaei2024persian, nazarizadeh2025parsoffensive}, or employ fragmented annotation schemes that capture only one or two aspects of the reasoning process, such as the target \cite{mozafari2024offensive} or the text span \cite{delbari2024spanning}. Thus, these approaches can identify whether content is hateful, but not {how it is expressed, why it is framed as harmful, or what potential consequences it may entail}. This limitation is particularly problematic because hate speech is often implicit, multi-layered, and morally framed, and cannot be reliably addressed using surface-level features alone \cite{kim2022hate}.

To bridge this gap, we propose an explainable multi-hop reasoning framework for hate speech detection in Persian. The sequential, human-interpretable framework supports a structured content-moderation process rather than relying on post-hoc explanations. Each hop builds on the previous one to reduce ambiguity and enable traceable and accountable decisions, capturing the underlying reasoning process by which humans identify, interpret, and assess hate speech.

Moral Foundations Theory (MFT) advances a pluralist view of morality grounded in multiple moral dimensions, including Care/Harm, Fairness/Cheating, Loyalty/Betrayal, Authority/Subversion, and Purity/Degradation \cite{koleva2015moral}. Hate speech often draws on shared moral values and perceived moral violations to legitimize hostility toward targeted groups \cite{hoover2019bound}. Given MFT's demonstrated relevance to Persian discourse \cite{atari2020foundations, nejat2015mental}, moral reasoning is crucial for understanding and detecting Persian hate speech.

By incorporating MFT, we construct a {1,200}-instance Persian hate speech dataset using a seven-hop reasoning framework: \textbf{(1) initial content triage}, \textbf{(2) target identification}, \textbf{(3) hate speech tactic identification}, \textbf{(4) moral foundation analysis}, \textbf{(5) textual rationale extraction}, \textbf{(6) inference of the underlying harmful narrative}, and \textbf{(7) severity and potential impact assessment}. Initial annotations are generated using {Gemini 3.5 Flash}, a state-of-the-art LLM in the Gemini family that has shown strong performance on Persian-language tasks,\footnote{\url{https://mcinext-mizan-llm-leaderboard.hf.space/}} and validated against independent human annotations. 


Using our dataset, we evaluate eight state-of-the-art proprietary and open-weight LLMs to investigate four questions: (1) how LLM performance changes as hate speech analysis progresses from initial triage to deeper reasoning; (2) how performance varies across proprietary versus open-weight and reasoning versus instruction-tuned models; (3) whether step-wise reasoning across Hops~1--6 improves severity prediction over one-shot prediction; and (4) which Persian hate speech tactics are most strongly associated with higher severity levels.

{Our findings reveal substantial variation in LLM performance across the multi-hop analysis of Persian hate speech. Performance generally declines as tasks require more fine-grained reasoning, with proprietary models generally outperforming open-weight models and the benefits of reasoning-oriented models varying across tasks. Incorporating intermediate analyses from Hops~1--6 substantially improves severity prediction over one-shot assessment, although the improvement varies considerably across models. Beyond model performance, \emph{Incitement to Violence or Harm} shows the strongest association with higher severity, while political-affiliation targets show a particularly strong association with {subversion}-based moral framing. These findings highlight the challenges of reliable, interpretable LLM-based hate speech analysis and the value of modeling harmful discourse for social media moderation.}

\section{Benchmark Construction}

To construct our benchmark, we consider two major social media platforms: X and Instagram, which are prominent sources of hate speech among Persian-speaking users. We first review recent studies that collect and release Persian-language datasets from these platforms (Appendix \ref{apx:dataSource}). Since most existing datasets primarily focus on hate speech instances, we balance the data at the first hop (initial content triage) by manually collecting recent posts from X. Specifically, we retrieve posts using keywords extracted from our labeled data by querying {Gemini 3.5 Flash}. Figures~\ref{fig:offensiveKeys}--\ref{fig:dangerousKeys} present example prompts used for extracting keywords associated with offensive language and dangerous speech. Overall, our dataset contains {348} instances collected from prior work and {852} instances collected in this study, resulting in {1,200} texts with {300} instances per initial content triage category. {The following describes the collection and preprocessing procedure for the newly collected instances.}

{To collect the 852 additional instances introduced in this benchmark, we manually search X using queries of the following format: \texttt{("keyword") lang:fa -is:retweet -is:reply since:2026-02-28}. This filtering strategy ensures that the collected posts were published after the collection period of existing datasets, which were all obtained before 2026, thereby reducing potential overlap and capturing more recent patterns of online discourse. We then apply the annotation prompts described in the following section to verify whether each retrieved post belongs to the target triage categories; posts that do not satisfy the criteria are removed. Finally, we anonymize all texts by replacing user mentions with \texttt{@username} and URLs with \texttt{www.link.com}. The overall workflow of our dataset construction and evaluation process is illustrated in} Figure~\ref{fig:pipeline}.

\paragraph{Data annotation.}


To assign values for each hop to every text instance, we employ {Gemini 3.5 Flash} with a chain-of-thought prompting strategy. The model is provided with the definition of each hop and its corresponding output schema, as specified in Table~\ref{tab:hopDef}. To assess the quality of the automatically generated annotations, we randomly select a sample of 100 texts, stratified by the first-hop content-triage labels (25 texts per category) to ensure balanced representation across the four categories. Two native Persian-speaking annotators independently annotate these texts using the same prompt provided to the LLMs. We then measure inter-annotator agreement between the two human annotators and compare the human annotations with the Gemini-generated reference annotations.

The inter-annotator agreement results indicate varying levels of agreement across annotation tasks and output types. Agreement is higher for initial content triage and severity classification, while it varies across categorical and multi-label annotations. Agreement is lower for tasks requiring open-ended judgments, particularly narrative inference and impact assessment. These results suggest that agreement depends on the annotation dimension and that text-based reasoning tasks involve greater interpretive variability among annotators. Complete inter-annotator agreement results, including the task-specific metrics used for each hop, are reported in Table \ref{tab:IAA} (Appendix \ref{apx:humanVal}).

Disagreements are resolved by the first author to produce the gold labels for each hop. We compare Gemini's outputs with these gold labels using the same metrics as in the LLM evaluation described in the following sections. Table~\ref{tab:alignment} (Appendix~\ref{apx:humanVal}) presents the Human--LLM alignment results. We observe high alignment for most closed-ended tasks, particularly target group identification and hate tactic and moral foundation detection. Alignment is lower for tasks involving open-ended text generation, especially narrative inference and potential impact assessment. This is expected, as automatic metrics may not fully capture semantic similarity, particularly in a low-resource language such as Persian. Despite these limitations, the results indicate that Gemini's annotations are sufficiently aligned with the gold labels to serve as reference annotations for our analyses.

\begin{table*}[t]
\centering
\scriptsize
\setlength{\tabcolsep}{4pt}
\begin{tabularx}{\textwidth}{lXX}
\toprule
\textbf{Hop} & \textbf{Definition} & \textbf{Output Schema} \\
\midrule
Initial Content Triage &
Determines whether the content is harmful and identifies its primary category. &
Labels: Offensive Language; Hate Speech; Dangerous Speech; Not Harmful. \\
\midrule
Target Group Identification &
Identifies the explicit or implicit target of the speech and the basis of identity targeted by the content. &
{Targeted Group (multi-label); Basis of Identity (multi-label). Labels provided in Table} \ref{tab:targetUnified}. \\
\midrule
Hate Tactic Identification &
Analyzes the rhetorical strategy used to express or legitimize hostility toward the target. &
Multi-label: Slurs/Derogatory Language; Dehumanization; Harmful Stereotyping; Incitement to Violence or Harm; Calls for Discrimination or Exclusion; Conspiracy Theories. \\
\midrule
Moral Foundation Analysis &
Identifies moral foundations invoked or violated to justify hostility, grounded in Moral Foundations Theory. &
Multi-label: Harm; Betrayal; Subversion; Degradation. \\
\midrule
Textual Evidence Extraction &
Extracts explicit textual evidence supporting the identified targets, tactics, and moral foundations. &
Target Span; Tactic Span; Moral Span. \\
\midrule
Harmful Narrative Inference &
Infers the implicit ideological or toxic narrative promoted by the content. &
Concise free-text narrative description. \\
\midrule
Severity/Impact Assessment &
Synthesizes prior hops to assess severity and potential real-world consequences. &
Severity Level (Low, Medium, High, Severe); Potential Impact (free text). \\
\bottomrule
\end{tabularx}
\caption{Definitions of the reasoning hops and their corresponding output schemas.}
\label{tab:hopDef}
\end{table*}

\begin{table}[t]
\centering
\scriptsize
\setlength{\tabcolsep}{3pt} 
\begin{tabularx}{\columnwidth}{lX}
\toprule
\textbf{Category} & \textbf{Standard Values} \\
\midrule
Targeted Group &
Iranian Officials \& Politicians; Iranian Security \& Military Forces; Domestic Political Opponents; Clerics \& Religious Leaders; Human Rights, Media \& Academic Activists; Migrants \& Ethnic/Religious Minorities; Foreign Adversaries; Iranian Women; Iranian Men; Artists, Actors \& Celebrities; Students, Children \& Adolescents; Companies, Organizations, \& Their Employees \\
\midrule
Basis of Identity &
Ethnicity/Nationality; Religion/Belief; Social Class; Gender; Political Affiliation; Migration Status; Occupation/Profession; Organizational/Institutional Affiliation; Group Membership; Social/Family Role; Personal Identity; Mental/Behavioral Traits; Educational/Academic Background; Humanity/Personhood \\
\bottomrule
\end{tabularx}
\caption{Categories for targeted groups and identity bases used in the study.}
\label{tab:targetUnified}
\end{table}

\section{Experimental Setup}

\paragraph{Models.}

We evaluate {eight} state-of-the-art multilingual LLMs on our benchmark: Grok 4.6, GPT-5.2, o3, and Claude Sonnet 4.6 as proprietary models, and DeepSeek R1, DeepSeek V3, Qwen3 30B A3B Thinking 2507, and Qwen3 30B A3B Instruct 2507 as open-weight models. This set enables comparisons across model access types and reasoning capabilities, with all models officially supporting Persian.


\paragraph{Prompting.}

We use API calls for all models. All prompts use one-shot prompting, with the multi-hop annotation prompt additionally employing a chain-of-thought strategy. We set the temperature to 0 and top-$p$ to 1, prioritizing accuracy over creativity. Each prompt is executed independently. Prompt templates and their English translations are provided in Appendix~\ref{apx:examplePrompts}.

\section{Main Results}

\subsection{Model Evaluation}

\paragraph{Hop 1: Initial content triage.}

As this is a single-label multiclass classification task with class-balanced reference data, we report macro-averaged precision, recall, and F1. As shown in Table~\ref{tab:hop1to6Res}, all proprietary models outperform the open-weight models, among which DeepSeek R1 performs best. Reasoning variants generally outperform their paired instruct variants, primarily through higher recall with little loss in precision. Overall, initial Persian hate-speech triage varies across model families and inference paradigms, with proprietary models leading and reasoning offering a recall advantage among paired open-weight models.


For Hops 2--6, we restrict the evaluation to instances that are annotated as \textit{Harmful} in the reference labels. This filtering excludes \textit{Not Harmful} cases and ensures that subsequent hops assess each model's ability to identify specific hate-speech attributes {on a common gold-standard set of harmful instances, making the results comparable.}

\begin{table*}[t]
\centering
\scriptsize
\setlength{\tabcolsep}{3pt}
\begin{tabular}{lcccccccc}
\toprule
\textbf{Metric} &
\textbf{Grok 4.6} & 
\textbf{GPT-5.2} & 
\textbf{o3} &
\textbf{Sonnet 4.6} &
\textbf{DeepSeek R1} &
\textbf{DeepSeek V3} &
\textbf{Qwen3 Think} &
\textbf{Qwen3 Inst}\\
\midrule
\multicolumn{9}{c}{\textbf{Hop 1: Initial Content Triage}} \\
\midrule
Macro Precision    & \textbf{82.7} & 82.6 & 79.0 & 79.5 & 77.0 & 76.7 & 67.8 & 69.6 \\
Macro Recall         & \textbf{83.0} & 80.4 & 77.7 & 79.2 & 75.0 & 68.6 & 66.7 & 52.0 \\
Macro F1              & \textbf{82.8} & 79.7 & 77.8 & 79.2 & 75.0 & 68.5 & 66.4 & 48.0 \\
\midrule
\multicolumn{9}{c}{\textbf{Hop 2: Targeted Group}} \\
\midrule
Micro F1     & \textbf{71.1} & 66.0 & 68.1 & 66.7 & 60.1 & 58.3 & 42.7 & 43.0 \\
Macro F1    & \textbf{65.6} & 63.6 & 63.4 & 56.4 & 41.7 & 22.9 & 27.4 & 20.3 \\
\midrule
\multicolumn{9}{c}{\textbf{Hop 2: Basis of Identity}} \\
\midrule
Micro F1    & \textbf{73.4} & 64.1 & 65.9 & 65.8 & 63.5 & 63.5 & 45.0 & 48.1 \\
Macro F1   & \textbf{59.7} & 51.5 & 49.8 & 42.5 & 36.7 & 35.2 & 28.2 & 13.3 \\
\midrule
\multicolumn{9}{c}{\textbf{Hop 3: Hate Tactic}} \\
\midrule
Micro F1        & 85.4 & \textbf{85.8} & 83.3 & 75.0 & 68.9 & 75.8 & 68.5 & 60.5 \\
Macro F1       & 75.4 & \textbf{75.6} & 65.5 & 53.1 & 43.1 & 35.9 & 47.0 & 6.9 \\
Adapted F1   & 95.4 & \textbf{97.7} & 96.2 & 77.3 & 89.4 & 91.9 & 82.9 & 88.6 \\
\midrule
\multicolumn{9}{c}{\textbf{Hop 4: Moral Foundation}} \\
\midrule
Micro F1       & \textbf{78.3} & \textbf{78.3} & 75.4 & 69.7 & 70.4 & 68.0 & 61.1 & 61.9 \\
Macro F1      & \textbf{66.8} & 38.2 & 57.6 & 50.3 & 30.7 & 48.8 & 41.2 & 46.7 \\
Adapted F1  & 89.0 & 88.3 & \textbf{89.8} & 71.2 & 85.5 & 81.4 & 73.6 & 78.1 \\
\midrule
\multicolumn{9}{c}{\textbf{Hop 5: Target Rationale}} \\
\midrule
Jaccard                                               & 71.4 & 64.6 & \textbf{72.4} & 54.6 & 62.5 & 60.8 & 52.9 & 45.2 \\
$\mathrm{Jaccard}_{\text{max}}$       & \textbf{81.2} & 77.8 & 78.7 & 64.0 & 70.6 & 68.3 & 56.1 & 55.4 \\
BERTScore                                          & 84.2 & 82.2 & \textbf{85.6} & 67.2 & 79.1 & 78.0 & 72.1 & 65.9 \\
$\mathrm{BERTScore}_{\text{max}}$  & \textbf{88.7} & 88.2 & 88.3 & 71.5 & 82.7 & 81.3 & 73.6 & 70.7 \\
\midrule
\multicolumn{9}{c}{\textbf{Hop 5: Tactic Rationale}} \\
\midrule
Jaccard                                               & 74.3 & 63.2 & \textbf{75.0} & 50.4 & 60.6 & 64.6 & 53.2 & 46.4 \\
$\mathrm{Jaccard}_{\text{max}}$       & \textbf{89.5} & 86.5 & 88.2 & 70.1 & 80.7 & 81.0 & 66.0 & 75.1 \\
BERTScore                                          & 85.5 & 80.4 & \textbf{87.2} & 63.6 & 77.2 & 87.3 & 71.0 & 69.0 \\
$\mathrm{BERTScore}_{\text{max}}$  & \textbf{92.9} & 92.4 & 92.8 & 74.0 & 87.2 & 87.3 & 77.2 & 84.4 \\
\midrule
\multicolumn{9}{c}{\textbf{Hop 5: Moral Rationale}} \\
\midrule
Jaccard                                               & \textbf{56.4} & 52.4 & 56.1 & 41.0 & 45.6 & 44.8 & 38.2 & 28.7 \\
$\mathrm{Jaccard}_{\text{max}}$       & \textbf{68.0} & 67.6 & 66.7 & 59.6 & 63.2 & 57.2 & 42.4 & 50.3 \\
BERTScore                                          & 74.3 & 71.2 & \textbf{75.4} & 57.9 & 68.3 & 64.4 & 60.5 & 54.2 \\
$\mathrm{BERTScore}_{\text{max}}$  & 80.2 & 79.0 & \textbf{80.9} & 68.0 & 77.4 & 70.6 & 62.5 & 66.3 \\
\midrule
\multicolumn{9}{c}{\textbf{Hop 6: Narrative Inference}} \\
\midrule
BERTScore F1      & \textbf{57.7} & 54.6 & 56.6 & 47.4 & 54.9 & 54.0 & 47.4 & 47.0 \\
Cosine Similarity & \textbf{74.5} & 72.2 & 72.9 & 63.0 & 71.4 & 68.4 & 60.3 & 61.3 \\
chrF++                & \textbf{27.5} & 25.2 & 23.8 & 25.1 & 26.0 & 25.0 & 19.5 & 22.1 \\
\bottomrule
\end{tabular}%
\caption{LLM performance across Hops~1--6 of the multi-hop Persian hate-speech reasoning benchmark. All metrics are reported on a 0--100 scale, with task-appropriate evaluation metrics used for each hop. Higher values indicate better performance.}
\label{tab:hop1to6Res}
\end{table*}

\paragraph{Hop 2: Target group identification.}

To assess whether LLMs can identify the target of hate speech and the identity dimension under attack, we report micro- and macro-averaged F1 for the Targeted Group and Basis of Identity tasks. Micro F1 captures overall performance, while macro F1 gives equal weight to each category and is more sensitive to performance on less frequent categories. Table~\ref{tab:hop1to6Res} shows that all proprietary models outperform the open-weight models. The larger micro--macro gaps for open-weight models suggest greater difficulty identifying less frequent categories among the 12 targeted groups (Table~\ref{tab:targetUnified}). DeepSeek R1 performs best among open-weight models in macro F1. Both DeepSeek R1 and Qwen3 Think also achieve higher macro F1 than their paired instruct variants, with only modest differences in micro F1, suggesting that reasoning may improve coverage of less frequent categories.


For Basis of Identity detection, proprietary models again outperform open-weight models on both metrics. The substantial micro--macro gaps indicate continued difficulty identifying less frequent categories among the 14 identity bases (Table~\ref{tab:targetUnified}). The effect of reasoning is task-dependent: Qwen3 Think achieves higher macro but lower micro F1 than Qwen3 Instruct, while DeepSeek R1 performs similarly to DeepSeek V3 in micro F1 and slightly better in macro F1. Overall, performance varies across model families and inference paradigms, with the largest differences emerging in coverage of less frequent categories.


\paragraph{Hop 3: Hate tactic identification.}

As this is a multi-label classification task, we report micro- and macro-averaged F1 and adapted F1~\cite{trager2025mftcxplain}, which measures whether a model identifies at least one hate tactic present in the text. As shown in Table~\ref{tab:hop1to6Res}, there is a substantial gap between micro and macro F1 score, which indicates that models perform better on frequent hate tactics than on less frequent ones, particularly among open-weight models.


The effect of reasoning is mixed: DeepSeek R1 achieves lower micro and macro F1 than DeepSeek V3, while Qwen3 Think outperforms Qwen3 Instruct on both metrics, particularly macro F1. Thus, reasoning may improve coverage of less frequent hate tactics in some models. Adapted F1 scores are generally high, indicating that most models identify at least one applicable hate tactic once harmful content has been established.

\paragraph{Hop 4: Moral foundation analysis.}

We next examine whether models can move beyond the hate tactics identified in Hop~3 to recognize the underlying moral foundations of harmful content. As this is a multi-label classification task, we report micro- and macro-averaged F1 and adapted F1. As illustrated in Table~\ref{tab:hop1to6Res}, compared with Hop~3, micro F1 and adapted F1 decrease for every model, indicating that moral foundation identification is more challenging overall. However, macro F1 increases for DeepSeek V3 and Qwen3 Instruct, suggesting that the difficulty of identifying less frequent categories varies across models.


The advantage of proprietary models is less uniform than in Hop~3: although Grok performs best overall, DeepSeek V3 achieves a higher macro F1 than GPT-5.2, indicating competitive coverage of less frequent moral foundations. The effect of reasoning is also mixed: DeepSeek R1 slightly outperforms DeepSeek V3 in micro F1 but performs substantially worse in macro F1, while Qwen3 Think achieves a slightly lower micro F1 than Qwen3 Instruct, which has the higher macro F1. Thus, reasoning provides no consistent advantage for moral foundation identification. Adapted F1 remains high (71.2--89.8\%), indicating that most models identify at least one relevant moral foundation even when they struggle to identify the full set of applicable foundations.

\paragraph{Hop 5: Textual evidence extraction.}

To assess whether LLMs can extract textual evidence supporting the target, tactic, and moral foundation analyses, we measure lexical overlap using Jaccard similarity~\cite{jaccard1901etude} and semantic similarity using BERTScore~\cite{zhang2019bertscore}. We use \textit{hazm}\footnote{\url{https://github.com/roshan-research/hazm}} for Persian tokenization and \textit{HooshvareLab/bert-base-parsbert-uncased}, pretrained on Persian~\cite{farahani2021parsbert}, for BERTScore. As each rationale may contain multiple textual spans, we report average pairwise and maximum similarity across reference spans, capturing overall alignment and the strongest individual match, respectively. Table~\ref{tab:hop1to6Res} presents the results for the three rationale types.


Tactic rationales achieve the highest similarity, followed by target rationales, while moral rationales are the most difficult to reproduce. This suggests that models more readily identify evidence associated with explicit hate tactics than evidence supporting more abstract moral foundations. BERTScore consistently exceeds Jaccard similarity, indicating that generated rationales often preserve semantic content despite lexical differences. Maximum similarity also consistently exceeds average similarity, suggesting that models can identify individual matching evidence spans even when their overall rationales differ from the references.

Proprietary models show the clearest advantage on moral rationales, while open-weight models such as DeepSeek R1 and V3 achieve relatively competitive scores on target and tactic rationales. Reasoning provides no uniform advantage across rationale types and model families, suggesting that its benefits in some classification hops do not necessarily transfer to textual evidence generation.

\paragraph{Hop 6: Harmful narrative inference.}

To evaluate the similarity between LLM-generated and reference harmful narratives, we compute three metrics for each reference--generation pair: BERTScore F1~\cite{zhang2019bertscore} with \textit{HooshvareLab/bert-base-parsbert-uncased} for token-level semantic similarity, cosine similarity~\cite{reimers2019sentence} using \textit{paraphrase-multilingual-mpnet-base-v2} for embedding-space similarity, and chrF++~\cite{popovic2017chrf++} for character $n$-gram overlap with word-order information. We report mean scores across all pairs in Table~\ref{tab:hop1to6Res}.


Proprietary models generally outperform open-weight models, although DeepSeek R1 and V3 achieve competitive semantic similarity scores. DeepSeek R1 attains a BERTScore F1 close to GPT-5.2 despite lower cosine similarity. Across all models, chrF++ scores are considerably lower than semantic similarity scores, indicating substantial character-level differences despite semantic overlap.

The comparison between reasoning and instruct variants reveals different patterns across model families. DeepSeek R1 outperforms DeepSeek V3 on all three metrics, whereas Qwen3 Think outperforms Qwen3 Instruct only on BERTScore F1, with lower cosine similarity and chrF++ scores. Thus, reasoning variants do not consistently improve similarity across model families and evaluation metrics.

\paragraph{Hop 7: Severity and impact assessment.}

We assess whether LLMs can determine the severity of harmful content and articulate its potential impact. For severity assessment, we calculate macro-averaged precision, recall, and F1 for this single-label, multi-class classification task. Because the four severity levels are ordinal (Low, Medium, High, Severe), we additionally report weighted Cohen's ($\kappa$)~\cite{cohen1968weighted}, excluding instances without a valid severity label from this metric but retaining them for the others. We also conduct a one-shot ablation in which models predict severity directly from the original text without access to the intermediate outputs from Hops~1--6, to examine their contribution to severity prediction (prompt template in Figure~\ref{fig:ablationPrompt}). Table~\ref{tab:hop7Res} reports results for both settings.


Multi-hop analysis improves macro F1 for five models (DeepSeek V3, DeepSeek R1, GPT-5.2, Qwen3 Instruct, and Sonnet), but decreases it for Grok, o3, and Qwen3 Think. However, weighted Cohen's ($\kappa$) is higher in the multi-hop setting for every model, suggesting better ordinal agreement on instances with valid predictions. These findings provide qualified support for RQ3: intermediate analyses improve ordinal agreement consistently but do not improve macro F1 for all models. The comparison between proprietary and open-weight models reveals no uniform advantage across settings, and reasoning variants do not consistently outperform their paired instruct variants: DeepSeek R1 underperforms DeepSeek V3 in macro F1 in both settings, while Qwen3 Think performs worse than Qwen3 Instruct in the multi-hop setting but better in the one-shot ablation.

For potential impact, we use BERTScore F1, cosine similarity, and chrF++, following Hop~6. GPT-5.2 achieves the highest scores on all three metrics. Across models, BERTScore and cosine similarity exceed chrF++, suggesting that generated statements capture the semantic content of the reference annotations despite lexical differences. Finally, in response to RQ1, severity prediction exhibits lower classification performance than initial content triage in Hop~1. Although both tasks involve four-class classification, Hop~7 requires a more fine-grained assessment of harmful content, suggesting that severity prediction is more challenging and supporting the broader observation that performance can decline as the benchmark progresses toward more complex reasoning tasks.

\begin{table*}[!htbp]
\centering
\scriptsize
\setlength{\tabcolsep}{3pt}
\begin{tabular}{lcccccccc}
\toprule
\textbf{Metric} &
\textbf{Grok 4.6} & 
\textbf{GPT-5.2} & 
\textbf{o3} &
\textbf{Sonnet 4.6} &
\textbf{DeepSeek R1} &
\textbf{DeepSeek V3} &
\textbf{Qwen3 Think} &
\textbf{Qwen3 Inst}\\
\midrule
\multicolumn{9}{c}{\textbf{Severity Level}} \\
\midrule
Macro Precision                     & 51.5 & 51.6 & 45.5 & \textbf{52.0} & 45.1 & 49.3 & 42.9 & 37.0 \\
Macro Recall                          & \textbf{53.1} & 51.7 & 47.7 & 40.5 & 34.4 & 43.0 & 31.9 & 33.5 \\
Macro F1                                & 50.7 & \textbf{50.8} & 44.6 & 43.5 & 32.1 & 43.9 & 27.9 & 29.9 \\
Weighted Cohen's $\kappa$ & \textbf{.829} & .819 & .760 & .797 & .558 & .736 & .525 & .525 \\
\midrule
\multicolumn{9}{c}{\textbf{Severity Level (Ablation)}} \\
\midrule
Macro Precision                      & \textbf{59.9} & 48.7 & 53.6 & 50.0 & 36.2 & 40.0 & 43.2 & 36.9 \\
Macro Recall                           & \textbf{60.3} & 46.6 & 54.6 & 45.5 & 30.9 & 33.5 & 37.1 & 37.2 \\
Macro F1                                & \textbf{58.4} & 42.8 & 53.1 & 43.4 & 21.8 & 27.0 & 31.8 & 29.2 \\
Weighted Cohen's $\kappa$ & \textbf{.731} & .615 & .702 & .595 & .254 & .336 & .301 & .259 \\
\midrule
\multicolumn{9}{c}{\textbf{Potential Impact}} \\
\midrule
BERTScore F1      & 59.1 & \textbf{61.4} & 59.6 & 50.2 & 58.5 & 57.4 & 53.2 & 53.2 \\
Cosine Similarity & 72.3 & \textbf{77.6} & 75.0 & 66.2 & 71.3 & 69.5 & 62.4 & 64.6 \\
chrF++                & 23.1 & \textbf{28.6} & 23.0 & 24.8 & 24.7 & 21.2 & 18.7 & 22.7 \\
\bottomrule
\end{tabular}%
\caption{LLM performance on hate speech severity (multi-hop and {one}-shot) and potential impact. All metrics are reported on a 0--100 scale, except for weighted Cohen's $\kappa$.}
\label{tab:hop7Res}
\end{table*}

\begin{figure*}[t]
  \centering
  \includegraphics[width=0.85\linewidth]{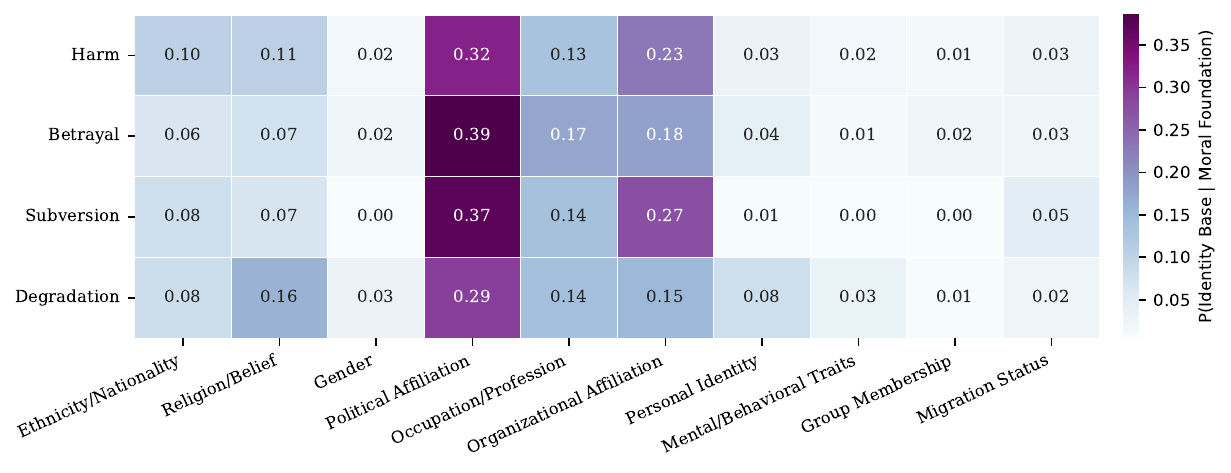}
  \caption{Conditional probability distribution of identity bases given the presence of each moral foundation.}
  \label{fig:identityMoral}
\end{figure*}

\subsection{Correlation Analysis Across Reasoning Hops}

\paragraph{Hate tactic vs. severity level.}

\begin{figure}[t]
  \centering
  \includegraphics[width=\columnwidth]{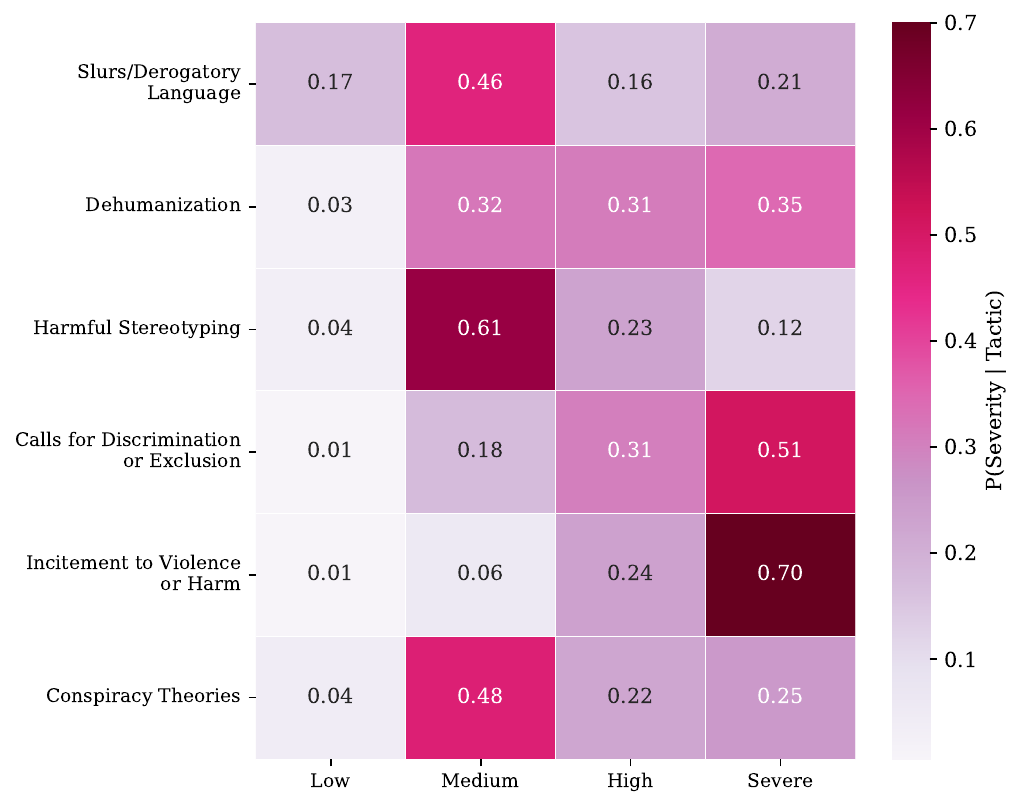}
  \caption{Conditional probability distribution of severity levels given the presence of each hate speech tactic.}
  \label{fig:tacticSeverity}
\end{figure}

To investigate which hate speech tactics in Persian are most strongly associated with higher severity levels (\textbf{RQ4}), we calculate the conditional probability of each severity level given the presence of a specific hate speech tactic in the reference data. Figure \ref{fig:tacticSeverity} {shows the resulting distributions and reveals a clear relationship between the type of tactic and severity. \emph{Incitement to Violence or Harm} is most strongly associated with the severe category (70\%), followed by \emph{Calls for Discrimination or Exclusion} (51\%). \emph{Dehumanization} also tends to occur at higher severity levels, with most instances falling into the medium or severe categories. In contrast, \emph{Harmful Stereotyping} and \emph{Conspiracy Theories} are predominantly associated with lower severity categories.}

To assess whether the observed differences are statistically significant, we conduct a chi-square test of independence separately for each tactic and apply Benjamini--Hochberg correction for multiple comparisons. The results, shown in Appendix \ref{apx:tacticSeverityStats}, {indicate significant associations between severity and five of the six tactics, with \emph{Conspiracy Theories} being the only exception. \emph{Incitement to Violence or Harm} shows the strongest association by far, followed by \emph{Slurs/Derogatory Language} and \emph{Calls for Discrimination or Exclusion}. This pattern is consistent with the distinction between direct and indirect forms of hateful expression, as conspiracy theories may operate more indirectly than tactics involving explicit harm, violence, or exclusion} \cite{Bilewicz01102024}.

\paragraph{Moral foundation vs. identity basis.}

We analyze co-occurrence patterns between moral foundations and identity bases to examine how identity-targeted hate speech is morally framed. Using the reference annotations, we compute the conditional distribution of identity bases for each moral foundation, excluding identity bases with fewer than 20 occurrences to reduce the influence of sparse categories. Figure~\ref{fig:identityMoral} presents the resulting distributions. {\emph{Political Affiliation} has high conditional probabilities across several moral foundations, with its strongest association with \emph{Subversion}, indicating that attacks targeting political affiliation are particularly likely to involve subversion-based moral framing.} {\emph{Organizational/Institutional Affiliation} and \emph{Occupation/Profession} also co-occur substantially with multiple moral foundations, highlighting the prominence of institutional and occupational identities in morally framed hate narratives. Less frequent identity bases show more limited and variable associations, which should be interpreted cautiously given their lower support.}

To test whether these patterns reflect systematic associations rather than differences in marginal frequencies, we conduct pairwise $\chi^2$ tests of independence using instance-level presence/absence indicators. We apply the Benjamini--Hochberg procedure to control the false discovery rate and report Cramér's $V$ as the effect size. Several associations remain significant after correction, {with the strongest effects for \emph{Harm} and \emph{Personal Identity} ($V=0.281$), \emph{Harm} and \emph{Organizational/Institutional Affiliation} ($V=0.273$), and \emph{Betrayal} and \emph{Political Affiliation} ($V=0.271$).} These results support the co-occurrence patterns observed in Figure~\ref{fig:identityMoral}. The complete set of significant associations is reported in Appendix~\ref{app:identityMoralStats}.

\paragraph{Content triage vs. targeted group.}

\begin{figure*}[htbp]
  \centering
  \includegraphics[width=0.7\textwidth]{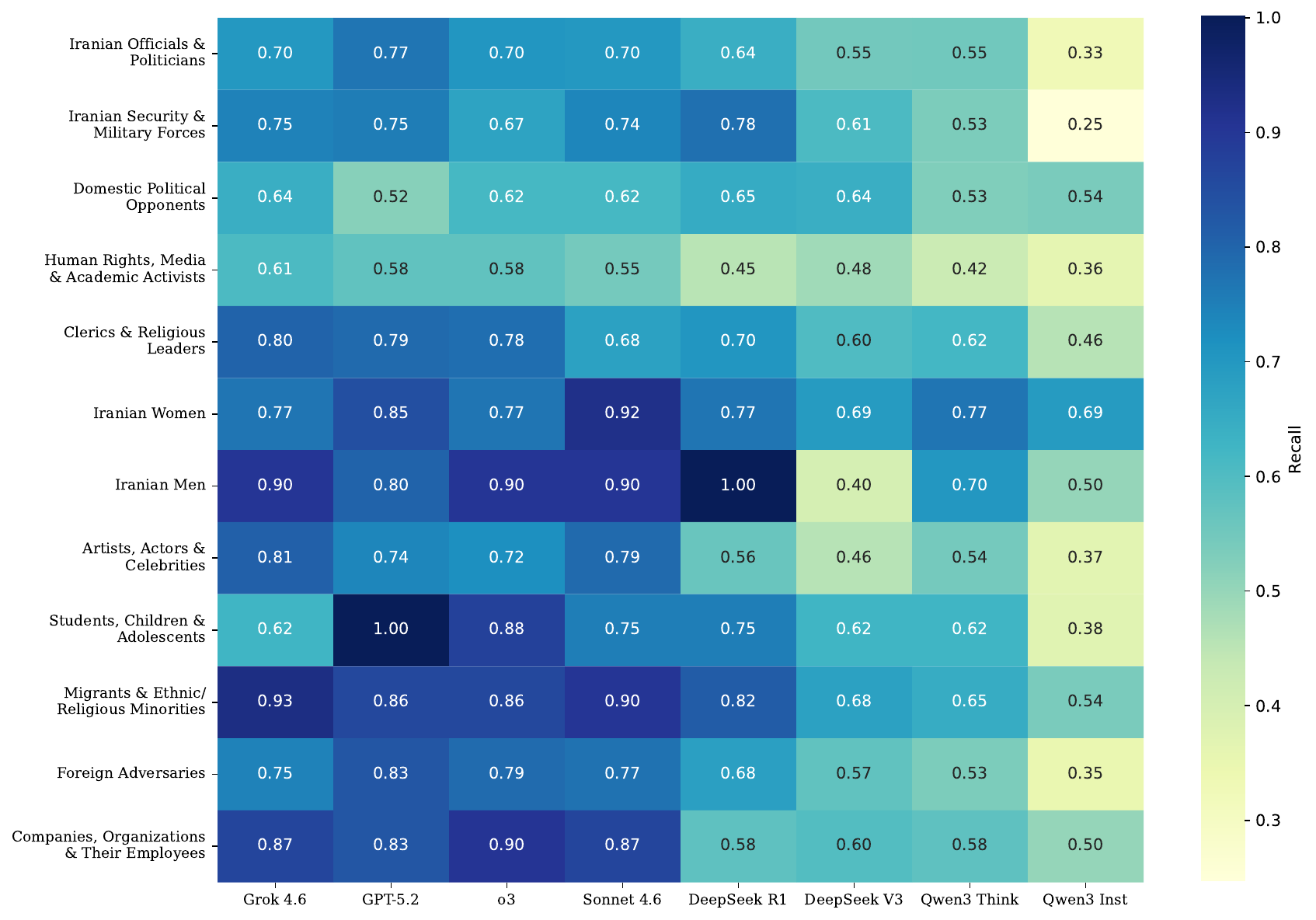}
  \caption{Recall of hate speech detection across targeted groups for each LLM.}
  \label{fig:groupHeat}
\end{figure*}

To further examine differences in LLMs' ability to identify specific targeted groups, we analyze recall across the 12 standardized categories. Figure~\ref{fig:groupHeat} presents the results. {\textit{Migrants \& Ethnic/Religious Minorities} is among the most consistently well-detected categories, with recall exceeding 80\% for Grok, GPT-5.2, o3, Claude, and DeepSeek R1. In contrast, \textit{Human Rights, Media \& Academic Activists} is consistently challenging, with recall below 61\% for every model and below 50\% for all open-weight models. Other categories, including \textit{Iranian Women}, \textit{Iranian Men}, and \textit{Companies, Organizations, \& Their Employees}, achieve relatively high recall for several models, although performance varies across model families. Reasoning variants generally improve recall: Qwen3 Think substantially outperforms Qwen3 Instruct on most target groups, while DeepSeek R1 generally outperforms DeepSeek V3, though not uniformly. Overall, performance varies considerably across target groups, with some categories consistently more challenging and reasoning-oriented models often providing broader coverage.}

\section{Conclusion}

In this work, we proposed a fine-grained multi-hop benchmark for analyzing Persian hate speech through increasingly complex reasoning steps. Our evaluation of state-of-the-art LLMs reveals substantial variation in performance across stages, with performance generally declining as tasks require more fine-grained semantic and inferential reasoning. Proprietary models generally outperform open-weight models, while the benefits of reasoning-oriented models vary across tasks. Our ablation study shows that intermediate analyses improve macro F1 for five of eight models, while weighted Cohen's $\kappa$ improves for all models, suggesting more consistent gains in ordinal agreement than in classification performance. Beyond model performance, our analyses uncover meaningful structure in Persian hate speech: \emph{Incitement to Violence or Harm} exhibits the strongest association with higher severity levels, while political-affiliation targets are particularly associated with subversion-based moral framing. Uneven performance across target groups also highlights differences in LLMs' sensitivity to hate speech directed at distinct social categories. Together, these findings demonstrate the challenges of reliable and interpretable LLM-based hate speech analysis and the value of modeling the underlying structure of harmful discourse, providing guidance for more effective and culturally informed moderation systems for Persian-language social media.




\section*{Limitations}

{Our study has several limitations that could be addressed in future work. We focus exclusively on hate speech in Persian and content from X and Instagram; extending our framework to other languages and platforms could provide a more comprehensive assessment of LLM performance and generalizability. Additionally, our benchmark is balanced across the four initial content-triage categories rather than reflecting their natural distribution on social media. Consequently, it may not represent the prevalence of different types of content in real-world settings. Future work could evaluate the framework on randomly sampled social media data to better reflect real-world distributions. Finally, the relatively small benchmark size may limit coverage of less frequent target groups, identity bases, and hate tactics. Larger and more diverse benchmarks could therefore provide broader coverage of these categories and enable more robust evaluation of model performance.}


\bibliography{custom}

\appendix

\section{Data Sources and Their Statistics}\label{apx:dataSource}

Table \ref{tab:dataStat} summarizes the datasets used in our study and their statistics

\begin{table}[!htbp]
\centering
\scriptsize
\begin{tabular}{lll}
\toprule
\textbf{Data Source} & \textbf{Platform} & \textbf{No. in Benchmark}\\
\midrule
\citeauthor{kebriaei2024persian} & X & {146} \\
\citeauthor{sabri2021emopars} & X & {25} \\
\citeauthor{elahimanesh2025emotion} & X & {46} \\
\citeauthor{davardoustdark} & Instagram & {81}  \\
\citeauthor{nazarizadeh2025parsoffensive} & Instagram & 50 \\
\bottomrule
\end{tabular}
\caption{Statistics of datasets used to construct our benchmark, including source, platform, and instance counts.}
\label{tab:dataStat}
\end{table}

\section{Human Annotation Protocol}\label{apx:annotation}

\subsection{Annotators}

\begin{itemize}

\item Number of annotators: 2

\item Native language: Persian
\item Annotation language: Persian
\item Role: Independent human validation of automatically generated annotations
\item Education: undergraduate students in Computer Engineering 
\item Prior annotation experience: basic NLP coursework 
\item Compensation: course-related research assistance

\end{itemize}

\subsection{Annotation Task}

For each text, annotators independently:

\begin{enumerate}
\item Assign an initial content-triage label.
\item Identify the targeted group(s) and basis(es) of identity.
\item Identify the hate speech tactic(s).
\item Identify the relevant moral foundation(s).
\item Extract textual evidence corresponding to the target, tactic, and moral foundation.
\item Describe the underlying harmful narrative.
\item Assign a severity level and describe the potential impact.
\end{enumerate}

\subsection{Key Instructions}

\begin{itemize}
\item Annotate each text independently without access to the other annotator's responses or the Gemini-generated annotations.
\item Follow the definitions and label sets provided in the annotation prompt.
\item For multi-label tasks, select all labels supported by the text.
\item For textual evidence extraction, select the shortest span that directly supports the corresponding annotation.
\item For narrative inference and potential impact, provide concise responses grounded in the content of the text.
\item Do not use external AI tools during annotation.
\end{itemize}

\section{Human Validation Results}\label{apx:humanVal}

The metric values for assessing inter-annotator agreement and human--LLM alignment are reported in Tables \ref{tab:IAA} and \ref{tab:alignment}, respectively.

\begin{table}[!htbp]
\centering
\scriptsize
\setlength{\tabcolsep}{5pt}
\begin{tabular}{ll}
\toprule
\textbf{Hop / Annotation Dimension} & \textbf{IAA Results} \\
\midrule
Hop 1: Initial Content Triage &
Cohen's $\kappa$: .743 \\
& Observed Agreement: 81.0 \\
\midrule
Hop 2: Targeted Group &
Micro F1: 82.0 \\
& Macro F1: 69.3 \\
\midrule
Hop 2: Basis of Identity &
Micro F1: 66.3 \\
& Macro F1: 51.2 \\
\midrule
Hop 3: Hate Tactic &
Micro F1: 74.9 \\
& Macro F1: 61.1 \\
\midrule
Hop 4: Moral Foundation &
Micro F1: 77.2 \\
& Macro F1: 60.6 \\
\midrule
Hop 5: Target Rationale &
Jaccard: 73.2 \\
\midrule
Hop 5: Tactic Rationale &
Jaccard: 80.3 \\
\midrule
Hop 5: Moral Rationale &
Jaccard: 76.1 \\
\midrule
Hop 6: Narrative Inference &
BERTScore-F1: 48.5 \\
& Cosine Similarity: 58.8 \\
& chrF++: 19.2 \\
\midrule
Hop 7: Severity Level &
Weighted Cohen's $\kappa$: .822 \\
& Observed Agreement: 61.8 \\
\midrule
Hop 7: Potential Impact &
BERTScore-F1: 53.7 \\
& Cosine Similarity: 66.4 \\
& chrF++: 24.3 \\
\bottomrule
\end{tabular}
\caption{Inter-annotator agreement results across annotation hops.}
\label{tab:IAA}
\end{table}

\begin{table}[!htbp]
\centering
\scriptsize
\setlength{\tabcolsep}{5pt}
\begin{tabular}{ll}
\toprule
\textbf{Hop / Annotation Dimension} & \textbf{Human--LLM Alignment Results} \\
\midrule
Hop 1: Initial Content Triage &
Macro Precision: 89.0 \\
& Macro Recall: 90.3 \\
& Macro F1: 88.8 \\
\midrule
Hop 2: Targeted Group &
Micro F1: 94.0 \\
& Macro F1: 94.3 \\
\midrule
Hop 2: Basis of Identity &
Micro F1: 88.2 \\
& Macro F1: 79.1 \\
\midrule
Hop 3: Hate Tactic &
Micro F1: 90.1 \\
& Macro F1: 80.6 \\
& Adapted F1: 97.4 \\
\midrule
Hop 4: Moral Foundation &
Micro F1: 92.0 \\
& Macro F1: 82.7 \\
& Adapted F1: 97.4 \\
\midrule
Hop 5: Target Rationale &
Jaccard: 86.5 \\
& $\mathrm{Jaccard}_{\text{max}}$: 91.8 \\
& BERTScore: 91.3 \\
& $\mathrm{BERTScore}_{\text{max}}$: 93.3 \\
\midrule
Hop 5: Tactic Rationale &
Jaccard: 84.3 \\
& $\mathrm{Jaccard}_{\text{max}}$: 92.7 \\
& BERTScore: 91.0 \\
& $\mathrm{BERTScore}_{\text{max}}$: 95.1 \\
\midrule
Hop 5: Moral Rationale &
Jaccard: 81.9 \\
& $\mathrm{Jaccard}_{\text{max}}$: 89.7 \\
& BERTScore: 89.2 \\
& $\mathrm{BERTScore}_{\text{max}}$: 93.2 \\
\midrule
Hop 6: Narrative Inference &
BERTScore-F1: 49.8 \\
& Cosine Similarity: 64.9 \\
& chrF++: 23.4 \\
\midrule
Hop 7: Severity Level &
Macro Precision: 72.7 \\
& Macro Recall: 67.3 \\
& Macro F1: 69.4 \\
& Weighted Cohen's $\kappa$: 95.5 \\
\midrule
Hop 7: Potential Impact &
BERTScore-F1: 55.0 \\
& Cosine Similarity: 68.1 \\
& chrF++: 26.2 \\
\bottomrule
\end{tabular}
\caption{Human--LLM alignment results across annotation hops.}
\label{tab:alignment}
\end{table}

\section{Prompt Templates}\label{apx:examplePrompts}

\subsection{Keyword Detection Prompts}

\begin{figure}[!htbp]
  \centering
  \includegraphics[width=\columnwidth]{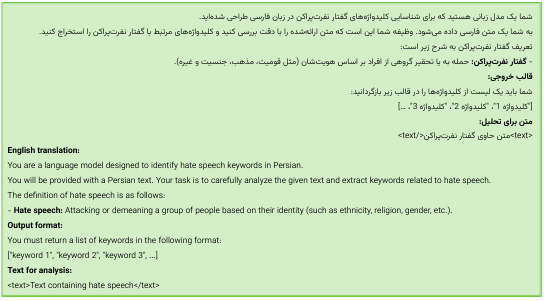}
  \caption{Prompt template for detecting hate speech keywords from a given text, along with its English translation.}
  \label{fig:hateSpeechKeys}
\end{figure}

The prompt templates used for hate speech keyword detection across the categories in initial content triage are provided in Figures~\ref{fig:hateSpeechKeys}--\ref{fig:hateSpeechKeys}, along with their English translations.

\begin{figure}[!htbp]
  \centering
  \includegraphics[width=\columnwidth]{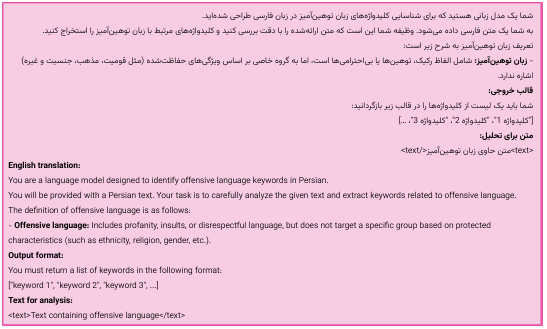}
  \caption{Prompt template for detecting offensive language keywords from a given text, along with its English translation.}
  \label{fig:offensiveKeys}
\end{figure}

\begin{figure}[!htbp]
  \centering
  \includegraphics[width=\columnwidth]{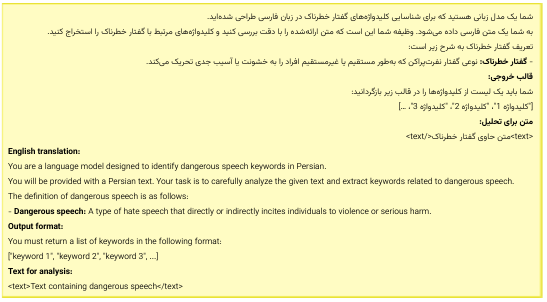}
  \caption{Prompt template for detecting dangerous speech keywords from a given text, along with its English translation.}
  \label{fig:dangerousKeys}
\end{figure}

\subsection{Ablation Study Prompt}

\begin{figure}[!htbp]
  \centering
  \includegraphics[width=\columnwidth]{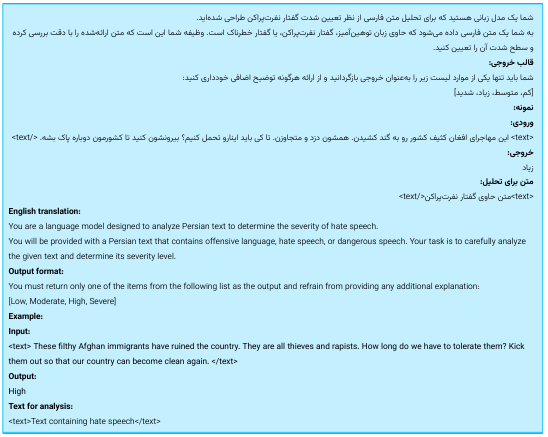}
  \caption{Prompt template for single-step severity classification of Persian hate speech in the ablation study, along with its English translation.}
  \label{fig:ablationPrompt}
\end{figure}

Figure~\ref{fig:ablationPrompt} presents the prompt template used for one-shot severity classification of Persian hate speech in the ablation study, along with its English translation.

\subsection{Multi-Hop Prompt}

\begin{figure*}[!htbp]
  \centering
  \includegraphics[width=\textwidth]{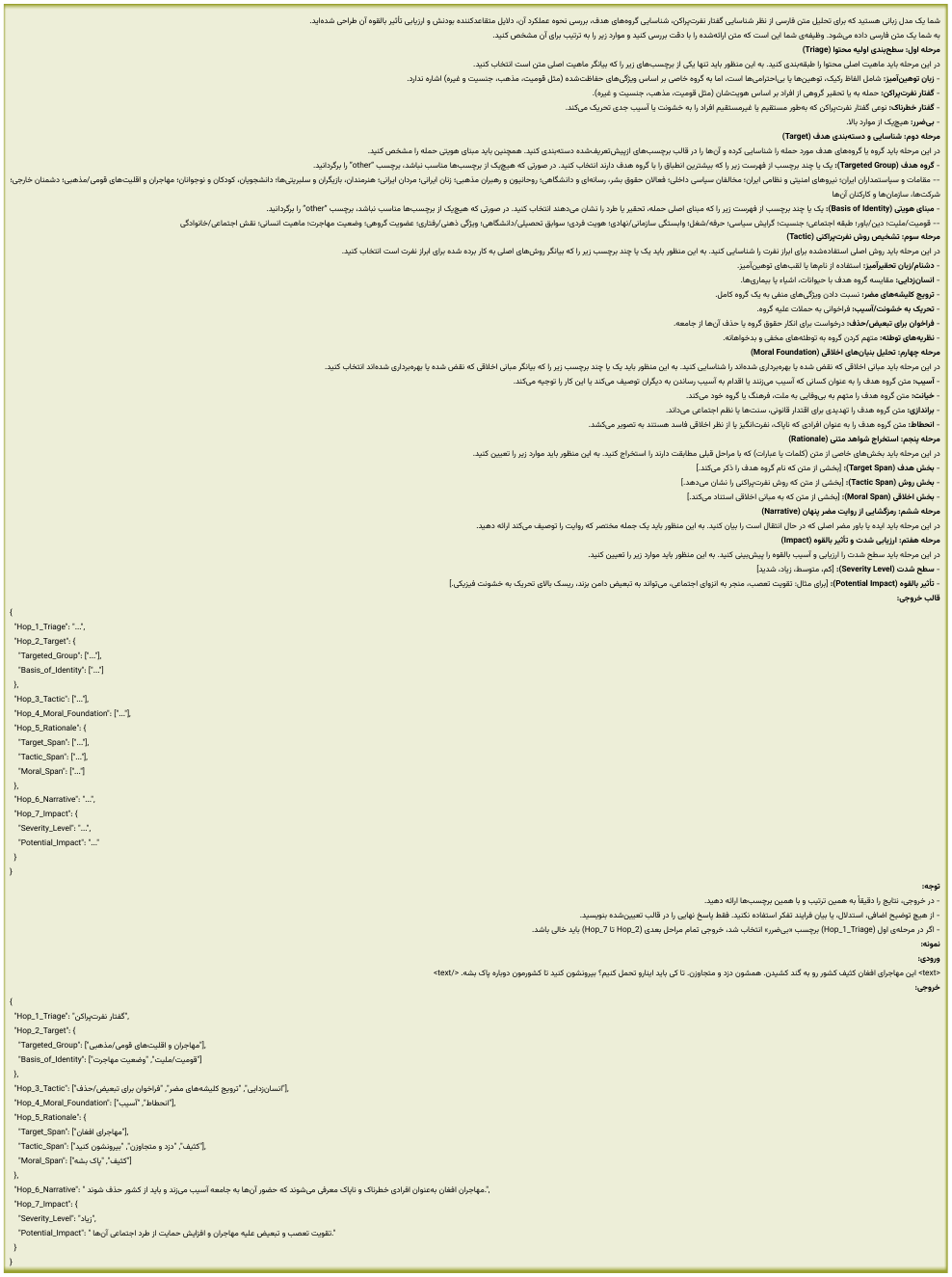}
  \caption{Prompt template for multi-hop analytical annotation of Persian hate speech.}
  \label{fig:annotatePrompt}
\end{figure*}

\begin{figure*}[!htbp]
  \centering
  \includegraphics[width=\textwidth]{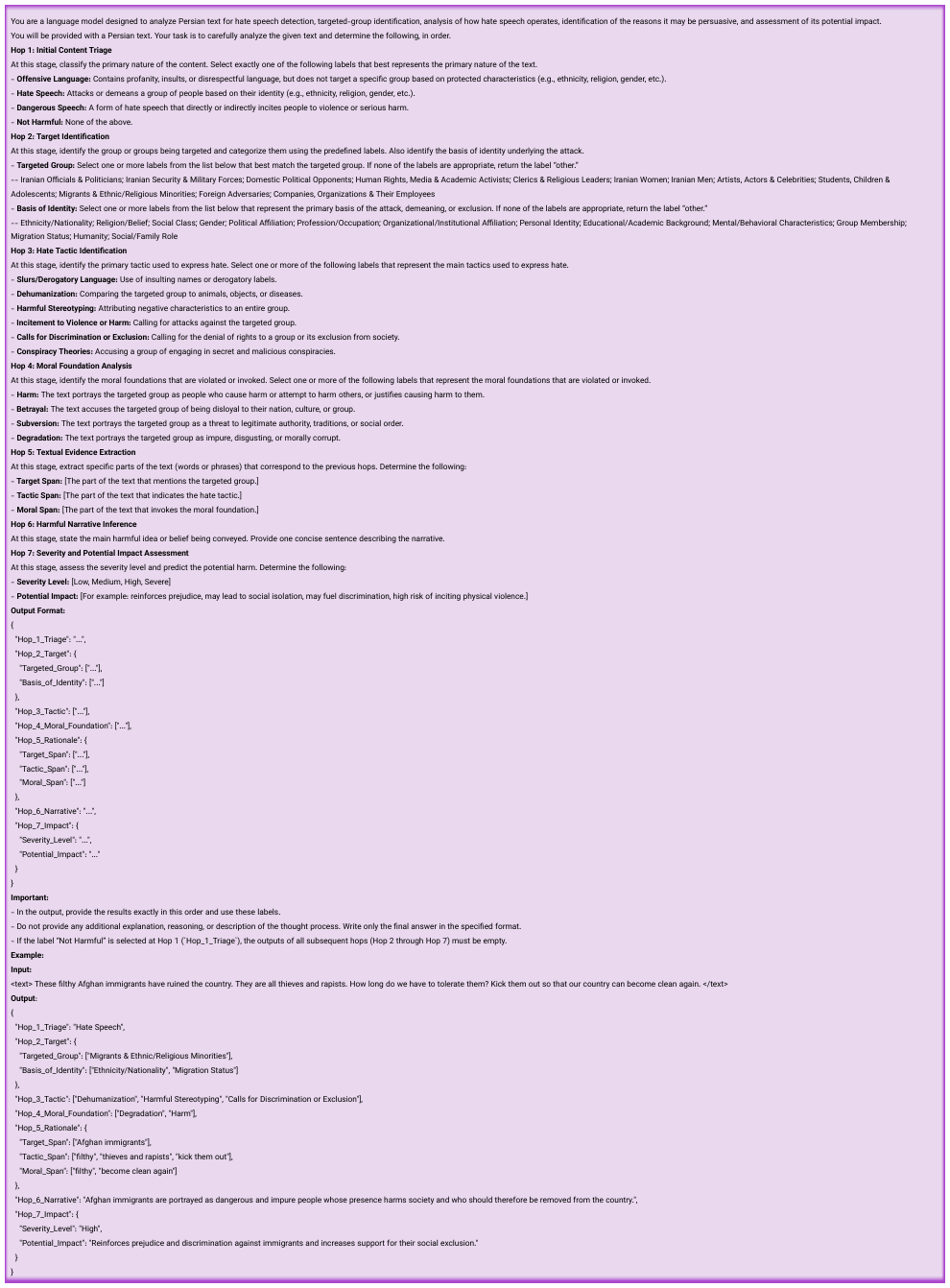}
  \caption{English translation of the prompt template for multi-hop analytical annotation of Persian hate speech.}
  \label{fig:annotatePromptEn}
\end{figure*}

The prompt template used for multi-hop analytical annotation of Persian hate speech and its English translation are presented in Figures \ref{fig:annotatePrompt} and~\ref{fig:annotatePromptEn}, respectively.

\section{Associations Between Hate Speech Tactics and Severity}\label{apx:tacticSeverityStats}

\begin{table}[t]
\centering
\resizebox{\columnwidth}{!}{%
\begin{tabular}{lrrrr}
\toprule
\textbf{Hate Speech Tactic} & $\boldsymbol{\chi^2}$ & \textbf{df} & 
$\boldsymbol{p_{\mathrm{adj}}}$ & \textbf{Cramer's $V$} \\
\midrule
Slurs/Derogatory Language
    & 118.79 & 3 & $<.001$ & .363 \\
Dehumanization
    & 63.48 & 3 & $<.001$ & .266 \\
Harmful Stereotyping
    & 71.53 & 3 & $<.001$ & .282 \\
Calls for Discrimination or Exclusion
    & 94.83 & 3 & $<.001$ & .325 \\
Incitement to Violence or Harm
    & 645.77 & 3 & $<.001$ & .847 \\
Conspiracy Theories
    & 6.96 & 3 & .073 & .088 \\
\bottomrule
\end{tabular}
}
\caption{Associations between hate speech tactics and severity levels. $p$-values are adjusted using the Benjamini--Hochberg procedure; Cramer's $V$ reports effect sizes.}
\label{tab:tacticSeverityStats}
\end{table}

\section{Statistical Analysis of Moral Foundation vs. Identity Basis Associations}\label{app:identityMoralStats}

\begin{table*}[!htbp]
\centering
\small
\begin{tabularx}{\textwidth}{lXrr}
\toprule
\textbf{Moral Foundation} & \textbf{Identity Basis} &
\textbf{$p_{\mathrm{BH}}$} & \textbf{Cramér's $V$} \\
\midrule
Betrayal
    & Political Affiliation
    & $<.001$ & .271 \\

Harm
    & Personal Identity
    & $<.001$ & .281 \\

Harm
    & Organizational/Institutional Affiliation
    & $<.001$ & .273 \\

Degradation
    & Religion/Belief
    & $<.001$ & .209 \\

Subversion
    & Organizational/Institutional Affiliation
    & $<.001$ & .199 \\

Degradation
    & Organizational/Institutional Affiliation
    & $<.001$ & .194 \\

Subversion
    & Political Affiliation
    & $<.001$ & .183 \\

Harm
    & Mental/Behavioral Traits
    & $<.001$ & .159 \\

Harm
    & Ethnicity/Nationality
    & $<.001$ & .149 \\

Betrayal
    & Occupation/Profession
    & $<.001$ & .147 \\

Degradation
    & Political Affiliation
    & .001 & .122 \\

Degradation
    & Gender
    & .001 & .122 \\

Betrayal
    & Religion/Belief
    & .001 & .121 \\

Degradation
    & Personal Identity
    & .001 & .121 \\

Subversion
    & Migration Status
    & .003 & .112 \\

Subversion
    & Personal Identity
    & .005 & .105 \\

Betrayal
    & Mental/Behavioral Traits
    & .007 & .100 \\

Betrayal
    & Group Membership
    & .025 & .086 \\

Harm
    & Political Affiliation
    & .047 & .078 \\
\bottomrule
\end{tabularx}
\caption{Significant associations between moral foundations and identity bases. Associations are tested using $\chi^2$ tests of independence based on instance-level presence/absence indicators. $p$-values are corrected using the Benjamini--Hochberg procedure. Cramér's $V$ is reported as the effect size.}
\label{tab:identityMoralStats}
\end{table*}

\end{document}